\documentclass[journal]{IEEEtran}
\ifCLASSINFOpdf
\else
\fi
\usepackage{graphicx}
\usepackage{amssymb}
\usepackage{amsmath}

\usepackage{xcolor}
\usepackage{multirow} 

\usepackage{textcomp}
\usepackage{algorithm}
\usepackage{algpseudocode}
\usepackage{cite}
\usepackage{hyperref}
\usepackage{caption}
\usepackage{CJKutf8}
\begin{document}
\begin{CJK}{UTF8}{gbsn}

\definecolor{darkgreen}{rgb}{0.0, 0.7, 0.0}

%
\title{FACT: Feature Adaptive Continual-learning Tracker for Multiple Object Tracking}
%
%
%


%
\author{Rongzihan Song,
        Zhenyu Weng, Huiping Zhuang, Jinchang Ren, Yongming Chen,
        and Zhiping Lin \thanks{Rongzihan Song, Zhenyu Weng, Yongming Chen, Zhiping Lin are with the School of Electrical and Electricity Engineering, Nanyang Technological University, Singapore. (e-mail: song0199@e.ntu.edu.sg, wzytumbler@gmail.com, yongming001@e.ntu.edu.sg, ezplin@ntu.edu.sg). (Corresponding author: Zhenyu Weng)}%
        \thanks{Huiping Zhuang is with the Shien-Ming Wu School of Intelligent Engineering, South China University of Technology, China. (e-mail: hpzhuang@scut.edu.cn)} %
        \thanks{Jinchang Ren is with the National Subsea Centre, Robert Gordon University, Aberdeen AB21 0BH, U.K. (e-mail: jinchang.ren@ieee.org).} } 
%

\markboth{ }%
{Shell \MakeLowercase{\textit{et al.}}: Bare Demo of IEEEtran.cls for IEEE Journals}
%



\maketitle

\begin{abstract}
Multiple object tracking (MOT) involves identifying multiple targets and assigning them corresponding IDs within a video sequence, where occlusions are often encountered. Recent methods address occlusions using appearance cues through online learning techniques to improve adaptivity or offline learning techniques to utilize temporal information from videos. However, most existing online learning-based MOT methods are unable to learn from all past tracking information to improve adaptivity on long-term occlusions while maintaining real-time tracking speed. On the other hand, temporal information-based offline learning methods maintain a long-term memory to store past tracking information, but this approach restricts them to use only local past information during tracking. To address these challenges, we propose a new MOT framework called the Feature Adaptive Continual-learning Tracker (FACT), which enables real-time tracking and feature learning for targets by utilizing all past tracking information. We demonstrate that the framework can be integrated with various state-of-the-art feature-based trackers, thereby improving their tracking ability. Specifically, we develop the feature adaptive continual-learning (FAC) module, a neural network that can be trained online to learn features adaptively using all past tracking information during tracking. Moreover, we also introduce a two-stage association module specifically designed for the proposed continual learning-based tracking. Extensive experiment results demonstrate that the proposed method achieves state-of-the-art online tracking performance on MOT17 and MOT20 benchmarks. The code will be released upon acceptance.



\end{abstract}

\begin{IEEEkeywords}
FACT, multiple object tracking, adaptivity, cascade association, continual learning.
\end{IEEEkeywords}\vspace{-10pt}

%
\IEEEpeerreviewmaketitle

\section{Introduction}
Online multiple object tracking (MOT) is a fundamental and challenging computer vision task. Aiming to solve the detection and tracking of multiple targets frame by frame, MOT has huge applications in safety monitoring, autonomous driving, etc \cite{zhang2022bytetrack, du2023strongsort}. According to the function of the MOT tracker, the task could be divided into two parts: object detection and target association \cite{cai2022memot}. Most of the recent researches focus on target association and many state-of-the-art methods have been developed based on the appearance and motion cues.  

Evidence suggests that appearance has a great capability for distinguishing targets in complex environments \cite{du2023strongsort}. This has led to a multitude of strategies prioritizing the optimization of appearance feature utilization. Some recent methods \cite{zhang2021fairmot, chu2019famnet, wojke2017simple} propose to utilize appearance cues using deep neural network-based (DNN-based) Re-Identification (ReID) models. These ReID models show enhanced discriminative abilities across various videos post-training on large-scale datasets compared to hand-crafted features \cite{sugimura2009using, yamaguchi2011you, izadinia20122}, contributing to re-association after losing the targets during tracking. Motivated by this, many recent methods focus on improving the discriminative ability of the ReID model with powerful siamese and transformer structures \cite{zhu2018distractor, zhang2019deeper,yu2022relationtrack} or optimizing the training efficiency \cite{yu2022towards, pang2021quasi}. However, despite these improvements, the discriminative capability remains limited in difficult tracking scenarios such as occlusions. Occlusions can lead to contaminated feature embeddings, as features might be sourced from the background or distractors.


To address occlusion problems, recent methods have started incorporating online learning techniques to improve adaptivity on appearance features or offline learning techniques to utilize temporal information from videos during tracking. Most existing online learning-based methods concentrate on learning from the recent frames to enhance their adaptivity to the latest appearance features \cite{xiang2015learning, zheng2021improving}. However, these approaches are unable to learn from all past tracking information to improve adaptivity on long-term occlusions while maintaining real-time tracking speed. To utilize the temporal information, recent offline learning methods \cite{cai2022memot, kim2021discriminative, gao2023memotr} propose the feature memory bank strategy that considers a set of features over a long period for tracking. However, these methods maintain a memory to store local past tracking information, limiting their ability to effectively utilize all past tracking information to enhance the discriminative capability of appearance features.

\begin{figure*}[tbp]
    \centering
    \includegraphics[width=0.90\textwidth,trim=30 235 95 82,clip]{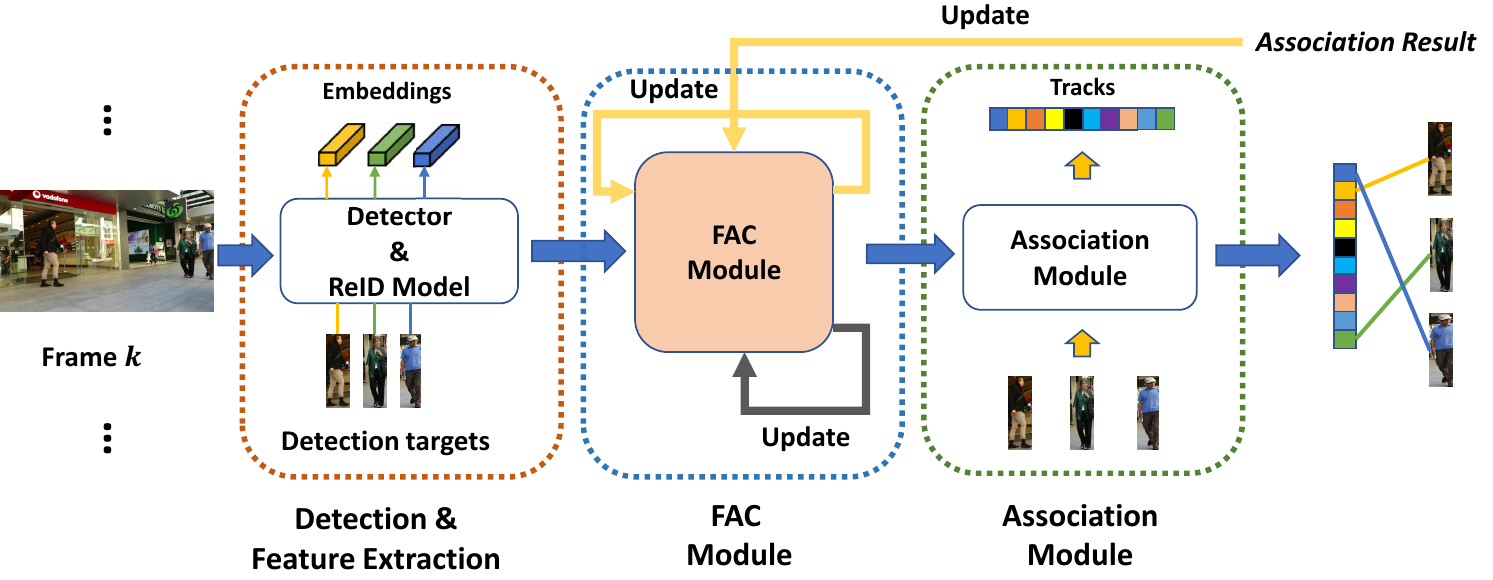}
    \caption{\textbf{Proposed FACT framework.} The FACT framework consists of three modules: the detection and feature extraction module, the FAC module, and the association module. Assume we are given any frame $k$ of a video, initially, the detection and feature extraction module identifies targets and converts their visual appearances into feature embeddings. Subsequently, these embeddings are processed by the FAC module. The output of the FAC module is then processed by the association module to associate the detected targets with their respective tracks, producing the association result. Finally, the FAC module performs online training using the embeddings, association results, and module parameters inherited from the last frame.}
    \label{fig:ADAPT Structure}
    \vspace{-17pt} 
\end{figure*}

To solve the above problems, this paper introduces a new MOT framework, the Feature Adaptive Continual-learning Tracker (FACT), that enables real-time tracking and feature learning for targets by utilizing all past tracking information, as illustrated in Fig. \ref{fig:ADAPT Structure}. We demonstrate that the framework can be integrated with various state-of-the-art feature-based trackers, thereby improving their tracking ability with minimal extra computation. To achieve this, we first introduce the Feature Adaptive Continual-learning (FAC) module, a neural network which can be trained online using all past tracking information simultaneously. To ensure efficient use of computational resources and minimize training time for real-time performance, we propose an analytic learning-based continual learning methodology to achieve the training recursively. In support of continual learning-based tracking, we also introduce a two-stage association module that begins with affinity estimation based on the FAC module's output, followed by instant association techniques such as cosine distance association. With the two-stage design, our association module is able to ensure robust tracking of new targets during the initialization of the FAC module. To assess the tracking performance of FACT, experiments were conducted on two benchmark datasets, MOT17 and MOT20. The results demonstrate that the proposed method achieves state-of-the-art online tracking performance in terms of HOTA.

Our contributions are summarized as follows: 

\begin{itemize}
\item [1)] We introduce a novel MOT framework, the Feature Adaptive Continual-learning Tracker (FACT), that enables real-time tracking and feature learning for targets by utilizing all past tracking information. We demonstrate that the framework can be integrated with various state-of-the-art feature-based trackers, thereby improving their tracking ability with minimal extra computation.

\item [2)] We introduce the Feature Adaptive Continual-learning (FAC) module, a neural network that employs the proposed analytic learning-based continual learning methodology to be efficiently and recursively trained online using all past tracking information simultaneously during tracking.

\item [3)] We present a two-stage association module specifically designed for the proposed continual learning-based tracker to ensure robust tracking of new targets during the initialization of the FAC module.\vspace{-7pt} 
\end{itemize}

\section{Related works}

\subsection{Tracking-by-detection Trackers}
The tracking-by-detection framework is the most popular framework to MOT challenges. Within this paradigm, targets are initially identified by a detector, followed by an association technique that links these targets to existing tracks. Thanks to the evolution of powerful detection models \cite{ge2021yolox, lin2017focal, zhou2019objects} and linear motion patterns during tracking, the motion-based method SORT \cite{bewley2016simple} simplified the association of targets to tracks using the efficient Intersection over Union (IoU) metric, resulting in fast processing. Subsequent research \cite{wojke2017simple, zhang2021fairmot} emphasized the role of target appearance features in tracking, thus many ReID model optimization-based methods have been proposed. DeepSORT \cite{wojke2017simple} integrated a ReID model to extract the appearance feature of targets to investigate appearance features for association. It shows that the ability of the ReID model to produce discriminative features is important for tracking. After this, many methods incorporated with advanced siamese \cite{zhu2018distractor, zhang2019deeper} and transformer \cite{yu2022relationtrack} structures to empower the ReID models or adopt contrastive learning \cite{yu2022towards, zhang2021fairmot, pang2021quasi} like triplet loss \cite{schroff2015facenet} to improve the discriminative capability by optimizing the ReID models training stage. However, all these methods focus on optimizing the ReID models themselves, and ignore the importance of updating the appearance features during tracking.\vspace{-10pt}

\subsection{Feature Update Approaches for MOT}
Appearance feature-based tracking relies on the comparison between the stored feature of tracks and the feature embeddings from targets. Thus, the feature updating strategy for stored features plays an important role in determining the overall tracking performance, particularly in handling occlusion scenarios. Several temporal information-based feature update methods have been proposed to enhance the adaptivity during the tracking process. Temporal information-based methods refine features to mitigate contamination from distractors by jointly considering features from multiple frames. For example, the weighted combination of the past and current track features has been widely adopted in many methods \cite{cao2023observation, zhang2021fairmot}. Later, MeMOT \cite{cai2022memot}, BLSTM-MTP \cite{kim2021discriminative} and MeMOTR \cite{gao2023memotr} propose the feature memory bank to maintain the long-term memory of past appearance data to update features more adaptively. However, while temporal information-based methods maintain a memory to store past tracking information, this approach restricts them to using only local past information, either because of hardware constraints on memory size \cite{cai2022memot, gao2023memotr} or because the memory network tends to focus more on recent data \cite{kim2021discriminative}. As a result, they cannot effectively utilize all past tracking information to enhance their discriminative capability of appearance features. To solve these problems, we employ continual learning techniques for online tracking, effectively utilizing all past tracking information.\vspace{-10pt}

\subsection{Continual Learning Approaches for Tracking}
Continual Learning (CL) aims to develop intelligent systems with the capability to continually adapt and learn from new tasks while retaining knowledge from previously encountered tasks. This technique enables the neural network to instantly update the knowledge from unseen data, thereby enhancing adaptivity and generalization \cite{marvasti2021deep, masana2022class}. To improve the adaptivity of appearance features, several online learning-based methods have been introduced. For instance, ref. \cite{bae2014robust} proposed using an Incremental Linear Discriminant Analysis (ILDA) classifier that learns from long-term past tracking information by training only on data from the latest frame. However, it shows limited performance due to the assumption that appearance features are linearly separable. Therefore, numerous neural network-based methods have been proposed to enhance discriminative capability. For instance, ref. \cite{bae2017confidence} proposed an online transfer learning approach using Convolutional Neural Networks (CNNs) with a fixed number of training samples. More recent approaches \cite{chu2017online, zheng2021improving} proposed to train separate SOT trackers online for each target using recent frames tracking information. However, SOT-based methods can be inefficient when handling a large number of targets. Additionally, to our knowledge, no existing online learning-based MOT method can learn from all past tracking information online while maintaining real-time tracking speed. To address these issues, we propose the Feature Adaptive Continual-learning (FAC) module, a unified neural network that can be trained online using all past tracking information. Furthermore, we propose an analytic learning-based methodology to achieve this training recursively, ensuring that processing speed remains consistent regardless of the video length, thereby maintaining real-time tracking performance. \vspace{-5pt}

\section{Methodology}

In this section, we will first revisit the popular MOT framework. Next, we will introduce the proposed FACT framework. Following that, we will provide a detailed explanation of our FAC module and the proposed analytic learning-based continual learning methodology. Finally, we will present the association module for target estimation.\vspace{-10pt} 





\subsection{Revisit the Popular MOT Framework}


The popular tracking-by-detection MOT framework is very straightforward. For every frame extracted from a video, this framework initiates by processing the image using pre-trained detection and ReID models. This step aids in identifying targets and subsequently converting their appearance feature into feature embeddings. After that, cosine distance is used to measure the dissimilarity between newly formed feature embeddings and stored track features for the association, followed by an IoU distance that utilizes the target's position information for further association. Finally, each track utilizes the association result to update the stored feature by weighted summation of previous and current feature embeddings. This procedure is reiterated for subsequent frames.\vspace{-10pt} 

\subsection{The Proposed FACT Framework}

The FACT framework consists of three main modules: the detection and feature extraction module, the FAC module, and the association module as illustrated in Fig. \ref{fig:ADAPT Structure}. Different from popular frameworks that solely utilize weighted summation feature update techniques for tracking, the distinguishing features of the FACT framework are its FAC module, which is trained online to continually learn target features, and the specially designed association module to associate targets with tracks. Subsequently, we illustrate the continual learning and association procedure of the FACT framework with the three modules.


Starting with frame $k$ (where $k=0,1, \cdots)$ from a video, the association procedure aligns with the popular framework by utilizing a pre-trained detector (specifically, YOLOX-X \cite{ge2021yolox}) and a Re-Identification (ReID) model (notably, FastReID SBS-50 \cite{he2020fastreid}). Their task is to convert the visual representation of the targets into one-dimensional feature embeddings. Thus, the feature embeddings in one frame can be represented as $\mathbf{X}_{k}^{\left (\textup{ReID}\right )}\in \mathbb{R}^{N_{k}\times d_{ReID}}$. In this context, $N_{k}$ denotes the number of targets, while $d_{ReID}$ stands for the embedding feature dimension. Subsequent to this extraction, the FAC module processes with these embeddings and passes the output to the association module for processing to generate the association results, associating these embeddings with the detected targets and the track identity set, symbolized as $T_{k} = \left \{ t_{0}, t_{1}, t_{2}...\right \}$. In the final, the FAC module leverages these association results, feature embeddings, and module parameters inherited from the last frame to learn target features online using the proposed continual learning methodology.\vspace{-5pt} 


\begin{table}[tbp]
\begin{center}
\caption{Notation Table For Reference In The Continual Learning Methodology.}
\label{Table:0}
\begin{tabular}{|c|c|}
\hline
\multirow{2}{*}{notation}                                           & \multirow{2}{*}{description}                                                                                                                    \\
                                                                    &                                                                                                                                                 \\ \hline
\multirow{2}{*}{$\mathbf{X}_{k}^{\left   (\textup{ReID}\right )}$}  & \multirow{2}{*}{feature embeddings at frame $k$}                                                                                                \\
                                                                    &                                                                                                                                                 \\ \hline
\multirow{2}{*}{$\mathbf{X}_{k}^{ \left (   \textup{et} \right )}$} & \multirow{2}{*}{feature embeddings after the transformation at frame $k$}                                                                        \\
                                                                    &                                                                                                                                                 \\ \hline
\multirow{2}{*}{$\mathbf{\hat{Y}}_{k}$}                             & \multirow{2}{*}{estimated labels at frame $k$}                                                                                                  \\
                                                                    &                                                                                                                                                 \\ \hline
$\mathbf{\hat{Y}}_{k}^{\left ( \textup{a}   \right )}$              & \begin{tabular}[c]{@{}c@{}}estimated labels of a-th group of tracks at frame $k$   \end{tabular} \\ \hline
$\hat{\mathbf{W}}_{\text{FCN}}^{\left (   \textup{k} \right )}$            & estimated FCN layer weight at frame $k$                                                                                                        \\ \hline
$ \mathbf{R}^{\left ( \textup{k} \right   )}$                       & feature autocorrelation unit at frame $k$                                                                                                       \\ \hline
\end{tabular}
\end{center}
\vspace{-15pt} 
\end{table}

\subsection{FAC Module}
In this subsection, we will discuss our continual learning methodology by detailing the architecture of the FAC module and the analytic learning-based methodology used to recursively learn target appearance features.

\begin{figure}[tbp]

    \centering
    \includegraphics[width=0.48\textwidth,trim=70 120 60 20,clip]{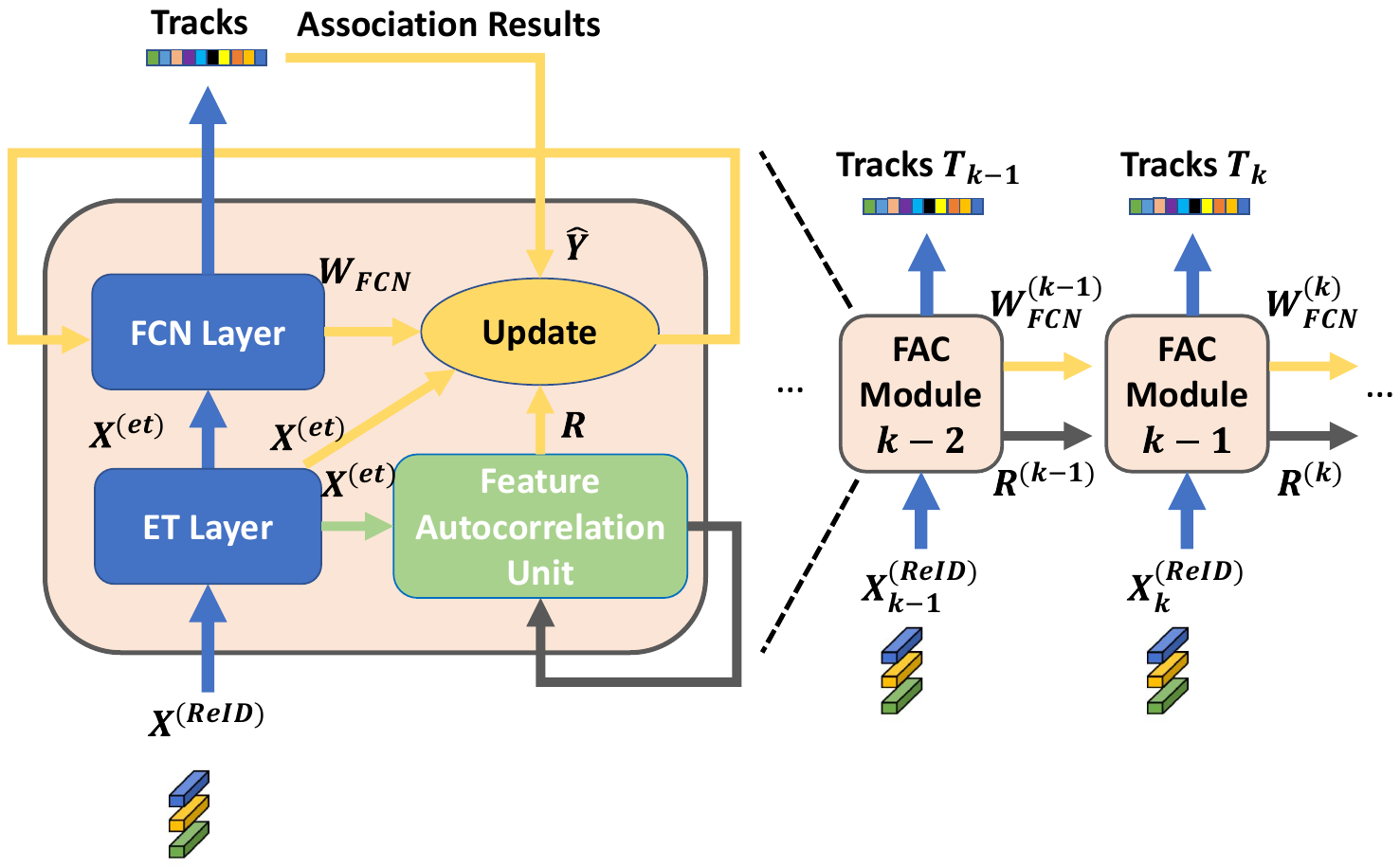}
    \caption{\textbf{The FAC module and online training process.} The FAC module consists of an embedding transformation (ET) layer and a fully-connected network (FCN) layer. Using the feature embeddings $\mathbf{X}_{k}^{\left (\textup{ReID}\right )}$ and corresponding association results from frame $k$, we update the FCN layer through online training. This training utilizes the feature embeddings, association results, FCN layer weight inherited from the last frame, and the feature autocorrelation unit, which is encrypted with both the current and past feature information.}
    \label{fig: ADAP Structure}
    \vspace{-10pt} 
\end{figure}

The architecture of the FAC module consists of an initial feature embedding transformation (ET) layer, succeeded by a learnable fully connected network (FCN) layer, acting as a classifier. This structure is depicted in Fig. \ref{fig: ADAP Structure}. For ease of reference, Table \ref{Table:0} provides an overview of the notations employed in our continual learning methodology.


The continual learning is realized by updating the FCN layer imitating the classification training process. To facilitate our discussion, we divide the continual learning into two phases: the base learning phase and the subsequent continual learning phase. We begin by introducing the base learning phase within the FAC module to build the foundation. Subsequently, we introduce the continual learning phase for the following frame utilizing the most recent data and association results to replicate the effects of base learning.


\subsubsection{Base Learning Phase} base learning phase is the core idea of the proposed continual learning methodology. It uses all the past appearance features for learning, enabling the network to efficiently utilize all past appearance information during tracking to improve discriminative capability. In the context of our continual learning, each target is uniquely associated with a single track, thus each track can be treated as a distinct class in classification training. 

Let $\left \{ \mathbf{X}_{k}^{\left (\textup{ReID}\right )} \right \}$ and the $\left \{ \mathbf{X}_{k}^{\left (\textup{ReID}\right )}, \mathbf{\hat{Y}}_{k} \right \}$ represents the feature samples and training samples at frame $k$. The symbol $\mathbf{\hat{Y}}_{k}\in \mathbb{R}^{N_{k}\times d_{T_{k}}}$ represents the corresponding association results, it works as the stacked one-hot labels for online training. Here, $d_{T_{k}}$ refers to the cumulative number of tracks present till frame $k$. It's worth noting that $\mathbf{\hat{Y}}_{k}$ is different from the ground truth (GT) labels, as they are obtained after the processing of each frame $k$. 


To contribute to our analysis, we divide the association results for each frame into different non-overlapping groups according to which frame the tracks first appear. For example, the association results for frame $k$ can be divided as $\mathbf{\hat{Y}}_{k}=\begin{bmatrix}
\mathbf{\hat{Y}}_{k}^{\left (0  \right )} & \mathbf{\hat{Y}}_{k}^{\left (1  \right )} &\cdots & \mathbf{\hat{Y}}_{k}^{\left (k  \right )}
\end{bmatrix}$, where each element $\mathbf{\hat{Y}}_{k}^{\left ( a \right )}\in \mathbb{R}^{N_{k}\times \left ( d_{ T_{a}}-d_{ T_{a-1}} \right )}$ represents the association results for a-th group of tracks at frame $k$. For $a=k$, the element $\mathbf{\hat{Y}}_{k}^{\left (k  \right )}$ represents the association results for new tracks that occur at frame k. If there is no new track, it will be set empty. We further simplify the association results as $\mathbf{\hat{Y}}_{k}=\begin{bmatrix}
\mathbf{\hat{Y}}_{k}^{\left (0:k-1  \right )}  & \mathbf{\hat{Y}}_{k}^{\left (k  \right )}
\end{bmatrix}$, where the $\mathbf{\hat{Y}}_{k}^{\left (0:k-1  \right )}\in \mathbb{R}^{N_{k}\times d_{ T_{k-1}} }$. Specifically, for frame $k=0$, the association results for the learning are obtained by manually assigning track identities to targets without any prior estimation.

Assume the base learning phase starts at frame $k-1$, given that $k\geq 1$, and we are given the feature embeddings $\mathbf{X}_{k-1}^{\left (\textup{ReID}\right )}$, the FAC module first applies an ET layer that transforms the input feature embedding into a higher-dimensional space follows most of the classifiers \cite{zhuang2022acil,guo2004pseudoinverse,zhuang2021correlation}. That is, the $\mathbf{X}_{k-1}^{\left (\textup{ReID}\right )}$ is transformed to $\mathbf{X}_{k-1}^{ \left ( \textup{et} \right )}$ as follows:\vspace{-5pt}
 
\begin{equation}
\label{eq:-1}
    \mathbf{X}_{k-1}^{\left ( \textup{et} \right )}= f_{\text{act}}\!\left(\mathbf{X}_{k-1}^{\left (\textup{ReID}\right )} \mathbf{W}_{\text{et}} \right ),
\end{equation}

\vspace{-5pt} In this context, $\mathbf{X}_{k-1}^{\left ( \textup{et} \right )} \in \mathbb{R}^{N_{k-1}\times d_{\text{et}}}$, with $d_{\text{et}}$ defining the transformation dimensionality. The matrix $\mathbf{W}_{\text{et}}$, which is of size $\mathbb{R}^{d_{\text{ReID}}\times d_{\text{et}}}$, serves as the transformation tool that modifies the ReID embeddings. Additionally, $f_{\text{act}}$ represents an activation function, with ReLU being our choice for this work. The elements of the matrix $\mathbf{W}_{\text{et}}$ were determined using randomization methods, aligning with the practices of the classifiers referenced earlier.




To effectively incorporate all past feature embeddings, the learning for frame $k-1$ is similar to training FCN layer weight using all past training samples $\left \{ \mathbf{X}_{i}^{\left (\textup{ReID}\right )}, \mathbf{\hat{Y}}_{i} \right \}_{i=0}^{k-1}$. As a result, the learning for frame $k-1$ can be approached by solving the following optimization problem:\vspace{-5pt} 
\begin{equation}
\label{eq:4}
\begin{array}{l}
\underset{\mathbf{W}_{\text{FCN}}^{\left ( k-1 \right )}}{\textup{argmin}}\left \|  \boldsymbol{H}_{k-1}
\right \|_{\textup{F}}^{2} + \gamma \left \| \mathbf{W}_{\text{FCN}}^{\left ( k-1 \right )} \right \|_{\textup{F}}^{2} 
\end{array},
\end{equation}
\vspace{-5pt} where $\gamma$ regularizes the above objective function, and 
\begin{equation}
\label{eq:4.1}
\boldsymbol{H}_{k-1} = \begin{bmatrix}
\mathbf{\hat{Y}}_{0}^{\left ( 0 \right )} & \mathbf{0}    & \cdots   & \mathbf{0}\\ 
\mathbf{\hat{Y}}_{1}^{\left ( 0 \right )} & \mathbf{\hat{Y}}_{1}^{\left ( 1 \right )}    & \cdots  & \mathbf{0}\\ 
\vdots &\vdots   &  \ddots  & \vdots\\ 
\mathbf{\hat{Y}}_{k-1}^{\left ( 0 \right )} & \mathbf{\hat{Y}}_{k-1}^{\left ( 1 \right )} & \cdots  & \mathbf{\hat{Y}}_{k-1}^{\left ( k-1 \right )}
\end{bmatrix} - \begin{bmatrix}
\mathbf{X}_{0}^{\left ( \text{et} \right )}\\ \mathbf{X}_{1}^{\left ( \text{et} \right )}
\\ \vdots 
\\ \mathbf{X}_{k-1}^{\left (\text{et} \right ) }
\end{bmatrix}
\mathbf{W}_{\text{FCN}}^{\left (k-1  \right )}.
\end{equation}

We simplify $\boldsymbol{H}_{k-1}$ to
\begin{equation}
\label{eq:4.2}
\boldsymbol{H}_{k-1} = \mathbf{\hat{Y}}_{k-1}^{ \text{history}} - \mathbf{X}_{k-1}^{ \text{history}}\mathbf{W}_{\text{FCN}}^{\left (k-1  \right )}.
\end{equation}
where $\mathbf{\hat{Y}}_{k-1}^{\text{history}}$ and $\mathbf{X}_{k-1}^{ \text{history}}$ represent all the past association results and transformed features until frame $k-1$, respectively.

The solution can be obtained as: \begin{equation}
\label{eq:6}
\begin{array}{l}
    \hat{\mathbf{W}}_{\text{FCN}}^{\left ( k-1 \right )} = \left ( \gamma \mathbf{I}+ \mathbf{X}_{k-1}^{\text{history}^T}\mathbf{X}_{k-1}^{\text{history}}\right )^{-1}\mathbf{X}_{k-1}^{\text{history}^T}\mathbf{\hat{Y}}_{k-1}^{\text{history}} ,
\end{array}
\end{equation}
where $\hat{\mathbf{W}}_{\text{FCN}}^{\left ( k-1 \right )}\in \mathbb{R}^{d_{\text{et}}\times d_{T_{k-1}}}$. It has a column size proportional to the existing tracks $d_{T_{k-1}}$. To this, we finalize the base learning for frame $k-1$.

To facilitate the analysis, we define a feature autocorrelation unit regards all the past feature embeddings up to frame $k-1$, as follows:
\begin{equation}
\label{eq:7}
\mathbf{R}^{\left ( \textup{k-1} \right )} = \left ( \gamma \mathbf{I}+ \mathbf{X}_{k-1}^{\text{history}^T}\mathbf{X}_{k-1}^{\text{history}}\right )^{-1}
\end{equation}
where $\mathbf{R}^{\left ( \textup{k-1} \right )}\in \mathbb{R}^{d_{\text{et}}\times d_{\text{et}}}$. Next, we will illustrate the continual learning phase for frame $k$.

\subsubsection{Continual Learning Phase} continual learning phase seeks to replicate the results of the base learning phase to the subsequent frame by employing only the latest tracking information. This is achieved with our proposed analytic learning-based methodology. This methodology can decrease both the time needed for learning and the resources required for storing all past appearance information during tracking.

Assuming we are processing frame $k$, the FCN layer weight $\mathbf{W}_{\text{FCN}}^{\left ( k \right )}$ for the base learning is obtained by solving the following:
\begin{equation}
\label{eq:9}
\begin{array}{l}
\underset{\mathbf{W}_{\text{FCN}}^{\left ( k \right )}}{argmin}\left \| \begin{bmatrix}
\mathbf{\hat{Y}}_{k-1}^{ \text{history}} & \mathbf{0} \\ 
\mathbf{\hat{Y}}_{k}^{ \left (0:k-1  \right )} & \mathbf{\hat{Y}}_{k}^{ \left (k  \right )}
\end{bmatrix}-\begin{bmatrix}
\mathbf{X}_{k-1}^{\text{history}}\\ 
\mathbf{X}_{k}^{\left ( \textup{et} \right )} 
\end{bmatrix}
\right \|_{\textup{F}}^{2} + \gamma \left \| \mathbf{W}_{\text{FCN}}^{\left ( k \right )} \right \|_{\textup{F}}^{2} 
\end{array}.
\end{equation}

In our continual leaning phase, we intend to obtain the identical FCN layer weight $\mathbf{W}_{\text{FCN}}^{\left ( k \right )}$ from base learning using only the $\hat{\mathbf{W}}_{\text{FCN}}^{\left ( k-1 \right )}$, feature correlation unit $\mathbf{R}^{\left ( \textup{k-1} \right )}$, transformed appearance feature $\mathbf{X}_{k}^{\left ( \textup{et} \right )}$ and the association results $\mathbf{\hat{Y}}_{k}$ from frame $k$. The solution to (7) using these components is summarized in Theorem 3.1.

\noindent\textbf{Theorem 3.1.} The solution to the FCN layer weight at frame $k$ could be obtained by
\begin{equation}
\label{eq:10}
\begin{array}{l}
\hat{\mathbf{W}}_{\text{FCN}}^{\left ( k \right )}  = \\  \begin{bmatrix}
\boldsymbol{V}_{k}\hat{\mathbf{W}}_{\text{FCN}}^{\left ( k-1 \right )}+ \mathbf{R}^{\left ( k \right )}\mathbf{X}_{k}^{\left ( \textup{et} \right )T}\boldsymbol{\hat{Y}}_{k}^{\left ( 0:k-1 \right )},   & \mathbf{R}^{\left ( k \right )}\boldsymbol{X}_{k}^{\left ( \textup{et} \right )\textup{T}}\boldsymbol{\hat{Y}}_{k}^{\left ( k \right )}
\end{bmatrix},
\end{array}
\end{equation}
where 
\begin{equation}
\label{eq:10.2}
\boldsymbol{V}_{k} = \left ( \mathbf{I}-\mathbf{R}^{\left ( k\right )}\mathbf{X}_{k}^{\left ( \textup{et} \right )T}\mathbf{X}_{k}^{\left ( \textup{et} \right )}  \right ),
\end{equation}

In this equation, the feature autocorrelation unit, denoted as $\mathbf{R}^{\left ( \textup{k}\right )}$, is updated as follows:
\begin{equation}
\label{eq:8}
\begin{array}{l}
\mathbf{R}^{\left ( k \right )}= \mathbf{R}^{\left ( k-1 \right )}\\-\mathbf{R}^{\left ( k-1 \right )}\mathbf{X}_{k}^{\left ( \textup{et} \right )T}(\mathbf{I}+\mathbf{X}_{k}^{\left ( \textup{et} \right )} \mathbf{R}^{\left ( k-1 \right )} \mathbf{X}_{k}^{\left ( \textup{et} \right )\textup{T}})^{-1} \mathbf{X}_{k}^{\left ( \textup{et} \right )}\mathbf{R}^{\left ( k-1 \right )},
\end{array}
\end{equation}

\emph{Proof.} {See the supplementary attached and also in the link:} \href{https://personal.ntu.edu.sg/ezplin/TMM-2023-Lin.pdf}{https://personal.ntu.edu.sg/ezplin/TMM-2023-Lin.pdf}.

With this, we have completed the continual learning phase. We have developed and validated an analytic learning-based methodology that replicates the identical results of the base learning phase for frame $k$ using only the latest tracking information. Consequently, the outcome for frame $k$ serves as the foundation for the continual learning phase in frame $k+1$, facilitating ongoing continual learning recursively. Together, the base learning and continual learning phases constitute our analytic learning-based continual learning methodology.

With the proposed continual learning approach, FACT can train the FAC module online during tracking. This allows FACT to achieve superior adaptivity to real-world scenarios that significantly differ from training data. Additionally, by training on all past appearance information during tracking simultaneously, the FAC module effectively utilizes all past appearance information to enhance discriminative capability.\vspace{-5pt}

\subsection{Association Module} In this subsection, we present the proposed two-stage association module designed for our continual learning-based tracker. This module is specifically designed to ensure robust tracking of new targets during the initialization of the FAC module. The association module operates in two stages: affinity estimation based on the FAC module's output and instant association with instant association techniques.

\subsubsection{Affinity Estimation} The aim of affinity estimation is to determine the feature affinity between the target's feature embeddings and the active tracks. These active tracks are the ones that remain in play and are relevant for tracking considerations.

Assume that we are dealing with the feature samples $\left \{ \mathbf{X}_{k}^{\left (\textup{ReID}\right )}  \right \}$ for frame $k$. We first use the FAC module to process these samples as follows:
\begin{equation}
\label{eq:11.2}
\boldsymbol{\hat{O}}{}_{k} =f_{\text{act}}\!\left ( \mathbf{X}_{k}^{\left (\textup{ReID}  \right )} \mathbf{W}_{\text{et}}\right )\hat{\mathbf{W}}_{\text{FCN}}^{\left ( \textup{k-1} \right )}
\end{equation}
Here, $\boldsymbol{\hat{O}}{}_{k} \in \mathbb{R}^{N_{k}\times d_{T_{k-1}}}$ represents the affinity between these feature embeddings and all the tracks. Subsequently, using $\boldsymbol{\hat{O}}{}_{k}$, the affinity values associated with the $M_{k}$ ($M_{k}\leqslant d_{T^{k}}$) active tracks  are extracted. This information is then used to construct the affinity matrix as follows:
\begin{equation}
\label{eq:12}
\boldsymbol{A}_{k} = \begin{array}{l}
\begin{bmatrix}
a_{0,0} & \cdots  & a_{0,M_{k}-1}\\ 
 \vdots & \ddots   & \vdots\\ 
a_{N_{k}-1,0} & \cdots & a_{N_{k}-1,M_{k}-1}
\end{bmatrix} \\
\end{array},
\end{equation}
where $a_{n,m}$ represent the affinity value between the $n$-th target and the $m$-th active track.  

It is noted that, unlike popular classification techniques, we do not apply a softmax operation to derive the affinity values. This strategy is designed based on the understanding that converting these values into classification probabilities could undermine their role as indicators of confidence. Without this strategy, even values indicating weak affinity might erroneously result in high-probability conclusions.

To make our analysis more straightforward, $\boldsymbol{A}_{k}$ is transformed into the affinity distance matrix $\boldsymbol{D}_{k}$, as follows:
\begin{equation}
\label{eq:13}
\boldsymbol{D}_{k} = \boldsymbol{1\cdot 1^{T}} - \boldsymbol{A}_{k}.
\end{equation}

The affinity distance matrix is combined with the Kalman filter to utilize motion information. This integration helps in associting targets with their corresponding tracks, a process which is efficiently achieved using the Hungarian algorithm\cite{kuhn1955hungarian}. Additionally, to enhance the accuracy of this association, a specific threshold is established for evaluating the affinity. For example, when targets are not fully initialized on the FAC module (meaning their confidence level is low), instant association techniques are employed to ensure the accuracy of the association process.

\subsubsection{Instant Association} In this stage, we adopt the most popular cosine distance for appearance feature measurement and Intersection over Union (IoU) distance for spatial measurement followed by the Hungarian algorithm for the instant association as in \cite{zhang2021fairmot, sun2019deep} to construct the FACT framework.

After the two-stage association, targets that remain unassociated with the tracks $T_{k}$ are considered as new tracks. Subsequently, the track set is updated to $T_{k}$, and the ReID appearance feature of each track is updated as in \cite{zhang2021fairmot}. To this, the estimation process is finished, producing the association results $\mathbf{\hat{Y}}_{k}$. These results serve both as the tracking output and the input for subsequent continual learning.

By employing the two-stage association module, the FACT framework benefits from both the exceptional discriminative capability of affinity estimation and the instant association strengths. Their combination leads to enhanced overall tracking performance.\vspace{-10pt}

\subsection{Computational Complexity Analysis}
As described above, the computational load of the proposed method mainly arises from the continual learning of the FAC module and the association process from the association module. For the continual learning of the FAC module, the complexity of feature autocorrelation unit updating is $T_{R} = O\!\left ( N_{k}^{3} + N_{k}d_{et}^{2}\right )$, where $N_{k}$ is the number of targets at frame $k$ and the $d_{et}$ is the transformed dimensionality of feature embeddings. The complexity of weight updating is $T_{W} = O\!\left ( d_{et}^{2} d_{T_{k}} + N_{k}d_{et}^{2}\right )$, where $d_{T_{k}}$ is the total number of track identities at frame $k$. Therefore, the combined complexity for the continual learning from the FAC module is $T_{R} + T_{W} = O\!\left ( N_{k}^{3} + N_{k}d_{et}^{2} + d_{et}^{2} d_{T_{k}}\right )$. For the association process from the association module, the complexity primarily arises from the affinity estimation. The complexity is $T_{estimation} = O\!\left ( N_{k} d_{T_{k}} \right )$ since the Hungarian algorithm is used to find the optimal number of associations. Therefore, the overall complexity of the proposed method is: \vspace{-10pt} 
\begin{align}
\label{eq:14}
T &= T_{R} + T_{W} + T_{estimation} & \notag \\
&= O\!\left ( N_{k}  d_{T_{k}} + N_{k}^{3} + N_{k}  d_{et}^{2} + d_{et}^{2} d_{T_{k}} \right ) 
\end{align}\vspace{-10pt}

In the contexts of the MOT17 and MOT20 scenarios, the dominant computational term is given by $d_{et}^{2} d_{T_{k}}$. Our analytic learning-based methodology significantly reduces the overall computational burden. As a result, the complexity of the computation increases linearly with the growth in the number of stored tracks.

\begin{table*}[tbp]
\begin{center}
\caption{Comparison between the association methods with and without the proposed FAC module. “w. FAC” is short for with FAC module, otherwise, we employ only the instant association techniques using naive cosine distance and IoU distance association. "Det" is short for the adopted detection model and "ReID" is short for the implemented ReID models. $\uparrow$ means higher is better, and $\downarrow$ means lower is better.}
\label{Table:1}
\begin{tabular}{l|ll|l|l|l|l|l|l}
\hline
\multicolumn{1}{c|}{\textbf{Dataset}} & \multicolumn{1}{c|}{\textbf{Det}} & \multicolumn{1}{c|}{\textbf{ReID}} & \multicolumn{1}{c|}{\textbf{Association}} & \multicolumn{1}{c|}{\textbf{HOTA $\uparrow$}} & \multicolumn{1}{c|}{\textbf{IDF1 $\uparrow$}} & \multicolumn{1}{c|}{\textbf{MOTA $\uparrow$}} & \multicolumn{1}{c|}{\textbf{IDSW} $\downarrow$} & {\textbf{FPS} $\uparrow$} \\ \hline
MOT17                                 & FairMOT                                & FairMOT                                 &                     &  57.3        &72.9                                & 69.2                               & 320                                & 26.1  \\
MOT17                                 & FairMOT                                & FairMOT                                 & w. FAC.             &  \textbf{58.1 \textcolor{darkgreen}{(+0.8)}}               & \textbf{74.0 \textcolor{darkgreen}{(+1.1)}}                               & 69.0                & \textbf{307}                       & 22.5  \\ \hline
MOT17                                 & FairMOT                                & FastReID                                 &                     & 58.6         & 74.9                               & 69.4                                & 276                                & 7.6  \\
MOT17                                 & FairMOT                                & FastReID                                 & w. FAC.            & \textbf{59.3 \textcolor{darkgreen}{(+0.7)}}                       & \textbf{75.9 \textcolor{darkgreen}{(+1.0)}}                                  & 69.3            & \textbf{274}                       & 7.1  \\ \hline
MOT17                                 & YOLO-X                                & FastReID                                 &                     & 69.2          & 81.8                               & 78.4                                & 157                                & 8.9  \\
MOT17                                 & YOLO-X                                & FastReID                                 & w. FAC.             & \textbf{69.9 \textcolor{darkgreen}{(+0.7)}}                   & \textbf{82.7 \textcolor{darkgreen}{(+0.9)}}                               & 78.3                & \textbf{165}                       & 7.1  \\ \hline
MOT20                                 & FairMOT                                & FairMOT                                 &                     & 65.0          & 82.2                               & 84.5                                & \textbf{2349}                               & 16.1  \\
MOT20                                 & FairMOT                                & FairMOT                                 & w. FAC.             & \textbf{65.4 \textcolor{darkgreen}{(+0.4)}}                   & \textbf{83.0 \textcolor{darkgreen}{(+0.8)}}                               & 84.5                & 2374                      & 13.0  \\ \hline
MOT20                                 & FairMOT                                & FastReID                                 &                    & 66.0           & 83.9                               & 84.6                                & 2096                               & 1.2  \\
MOT20                                 & FairMOT                                & FastReID                                 & w. FAC.            & \textbf{66.5 \textcolor{darkgreen}{(+0.5)}}                    & \textbf{84.8 \textcolor{darkgreen}{(+1.0)}}                              & 84.6                 & \textbf{2015}                      & 1.2  \\ \hline
MOT20                                 & YOLO-X                                & FastReID                                &                      & 68.9         & 84.2                              & 86.9                                 & \textbf{1140}                               & 2.1  \\
MOT20                                 & YOLO-X                                & FastReID                                 & w. FAC.             & \textbf{70.1 \textcolor{darkgreen}{(+1.2)}}                    & \textbf{86.2 \textcolor{darkgreen}{(+2.0)}}                               & 86.9                & 1146                      & 1.8  \\ \hline
\end{tabular}
\end{center}
\vspace{-20pt} 
\end{table*}

\section{Experiments}
\subsection{Settings}

\noindent\textbf{Datasets.} We evaluate the FACT on MOT17 \cite{milan2016mot16} and MOT20 \cite{dendorfer2020mot20} benchmarks under the “private detection” protocol. Both datasets provide the training set and testing set respectively without the validation set. For ablation studies on MOT17 and MOT20, we follow the \cite{zhou2020tracking, zhang2022bytetrack} to use the first half of each video in the training set for training, the second half for the validation. MOT17 mainly focuses on tracking under moving and static camera scenarios, and MOT20 focuses more on tracking under crowded environments. Two datasets together verify the performance of the FACT.

\noindent\textbf{Metrics.} We use the common CLEAR \cite{bernardin2008evaluating} metrics to evaluate the different aspects of the FACT tracking performance, including Higher-Order Tracking Accuracy (HOTA) \cite{luiten2021hota}, IDF1 \cite{ristani2016performance}, Multiple Object Tracking Accuracy (MOTA), ID Switch (IDSW), False Positive (FP), False Negative (FN), Frame Per Second (FPS). MOTA mainly focuses on object detection performance, and IDF1 focuses more on tracker association performance. HOTA is a recently proposed new metric that balances the performance of detection, association, and localization. The FACT aims to improve the association part. Therefore, we focus on IDF1 and HOTA values in this work.

\noindent\textbf{Implementation Details.} For the FACT, we follows the \cite{zhang2022bytetrack} to chooses the YOLOX as the default detector. To make a fair comparison, the training schedule remains the same to \cite{zhang2022bytetrack}. We choose FastReID's \cite{he2020fastreid} SBS-50 model as the default feature extractor, following the same training strategy to \cite{aharon2022bot}. The ET layer transformation size is chosen as 3000. As the standard practice, the simple camera motion compensation (CMC) technique is adopted in our method and kept the same to \cite{aharon2022bot}. In the linear assignment step of the association module, we adopt the cosine distance and IoU distance association as the instant association techniques to construct the FACT. As a common practice, for the final benchmarks, we apply simple linear interpolation, following the approach in \cite{zhang2022bytetrack}, unless otherwise specified.

For the ablation study, we adopt a different combination of detection and ReID models to evaluate the generalization ability of the FACT following the \cite{zhang2022robust}. We choose three different models, FairMOT, YOLO-X, and SBS-50 model for the experiments. Furthermore, we use the first half of the training data for training and the remaining for validation. The experiments are implemented using Pytorch and run on a server with NVIDIA GeForce RTX 2080Ti GPU unless otherwise specified. Following \cite{ge2021yolox}, the FPS is measured with the batch size of 1 on a single GPU under the FP16-precision.\vspace{-10pt}


\subsection{Evaluation of the FACT}

In this section, detailed experiments are presented to validate the advantages of FACT stated in the previous sections. Given that FACT primarily focuses on facilitating the association, emphasis is mainly placed on the HOTA and IDF1 score. \vspace{-15pt} 

Comparison of distances in subsequent frames between the FAC module (continual learning-based) and the commonly used cosine distance association (without continual learning) for selected identities.

\vspace{\baselineskip}
\noindent\textbf{Analysis on FAC module.} This section demonstrates that the FAC module, utilizing our proposed continual learning approach, has superior adaptivity on appearance features compared to methods that do not employ continual learning and instead rely solely on second-stage instant association with naive cosine and IoU distance metrics. To independently evaluate the effectiveness of the FAC module (rather than benefiting from other modules), we evaluate the tracking performance of methods that use different detection models, ReID models, and association approaches. The comparison is illustrated in Table \ref{Table:1}.

The results demonstrate that our FAC module significantly improves the IDF1 score, with an increase of approximately 1.1 on the MOT17 dataset and 1.3 on the MOT20 dataset compared to naive cosine distance and IoU distance associations. Similarly, HOTA values improved across all combinations, with gains of up to 0.8 on MOT17 and 1.2 on MOT20. Additionally, incorporating our FAC module with these naive association techniques resulted in a substantial reduction in ID switches. 

\begin{figure}[tbp]
    \centering
    \includegraphics[width=0.5\textwidth,trim=70 250 200 120,clip]{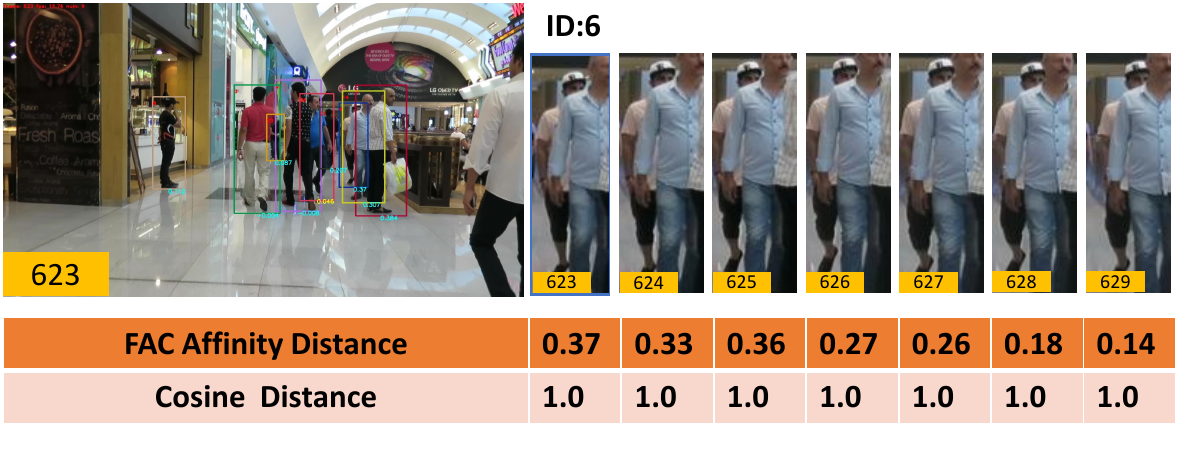}
    \caption{\textbf{Comparison of occlusion adaptation: FAC module-based affinity estimation vs. cosine distance.} \textbf{The top left large image:} Tracking results for frame 623, showing tracked objects in various colors with FAC module. \textbf{The top right smaller image sequence:} The occluded target image in the subsequent frames. \textbf{The bottom table:} Comparison of the distances in subsequent frames between the FAC module (continual learning-based) and the popular cosine distance association (without continual learning) for selected identities. The focus is on ID: 6, the individual dressed in pink who is partially occluded and positioned behind others. Our method successfully tracks the partially occluded target, whereas the cosine distance does not.
    }
    \label{fig: adaptivity}
    \vspace{-10pt} 
\end{figure} 

The consistent performance gain across all combinations can be attributed to the improved adaptivity brought by training the FAC module online during tracking. The larger improvement on the MOT20 dataset can be attributed to the improved discriminative capability of appearance features enabled by our utilization of all past tracking information during tracking. Given that MOT20 has much longer average video sequences (2233 frames per video) compared to MOT17 (759 frames per video), the benefits of using all past tracking information are greater.

\begin{figure}[tbp]
    \centering
    \includegraphics[width=0.5\textwidth,trim=95 90 250 300,clip]{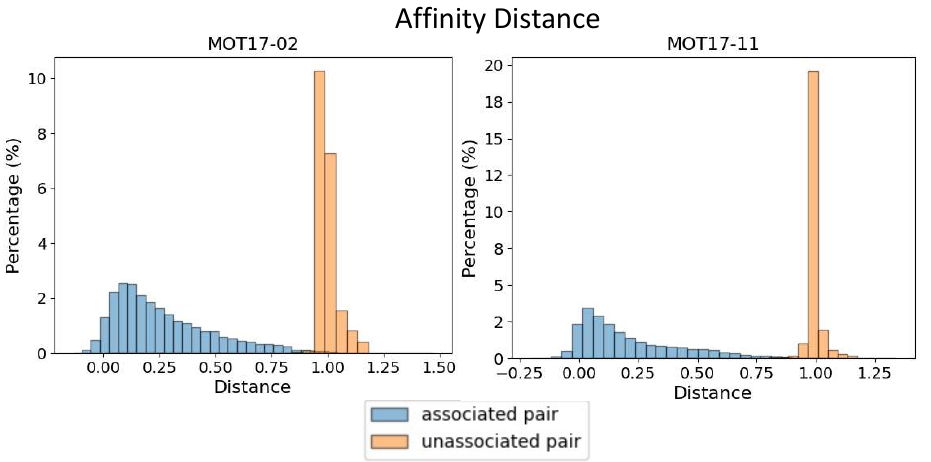}
    \caption{\textbf{Distance distribution of the associated and unassociated pairs on two videos from MOT17.} }
    \label{fig: ADAP distance}
    \vspace{-10pt} 
\end{figure} 

The FPS results demonstrate the computational efficiency of the FAC module. For larger models (YOLO-X + FastReID), the FPS decrease is minor, with a drop of 0.3 for MOT20 and 0.4 for MOT17. The FPS drop increases slightly with lighter models. On (FairMOT + FairMOT), the drop is 3.1 for MOT20 and 3.6 for MOT17. However, the FAC module’s computational cost is modest relative to its performance gains. For example, in MOT17, integrating the FAC module with (FairMOT + FairMOT) improves the IDF1 score to 74.0, which is close to the 74.9 achieved using the larger model combination (FairMOT + FastReID). Concurrently, the FAC module only reduces FPS by 3.6, a much smaller reduction compared to the 18.5 decrease when using the larger FastReID model. These findings highlight the FAC module’s efficiency.\vspace{-5pt}

\vspace{\baselineskip}
\noindent\textbf{Case Study1: FAC Module's Adaptivity to Occlusions.}
In this section, we evaluate the adaptivity of the FAC module, which employs a continual learning approach for effective MOT, particularly in handling frequent occlusions. As depicted in Fig. \ref{fig: adaptivity}, we compare the occlusion adaption using the affinity estimation with the FAC module against using the cosine distance for the target (ID: 6). As seems, the cosine distance fails to associate any tracks with the target due to heavy occlusion, resulting in a distance around 1.0 (we set an upper limit of 1.0 to offset the precision error). In contrast, the proposed FAC module, employing our continual learning approach, enhances adaptation to occlusions. Once associated at frame 623, the distance continues to decrease from 0.37 to 0.14 at frame 629, indicating a progressively improved feature reliability with the continual learning process.\vspace{-10pt}

\vspace{\baselineskip}
\noindent\textbf{Case Study2: Visualization of the FAC Module's Enhancement of Discriminative Capability.} To illustrate how the FAC module improves discriminative capability, we plotted the distance distribution between targets and tracks during the affinity estimation process, as shown in Figure \ref{fig: ADAP distance}. The unassociated pairs within our proposed association module typically exhibit distance values close to 1.0. This demonstrates that the FAC module effectively distinguishes unassociated pairs, validating the efficacy of our FACT approach. Additionally, the associated pairs show greater variability within the association module. This variability can be attributed to the FAC module requiring several frames to initialize each new track. If a new track isn't fully initialized, the confidence in FAC predictions may fall below a specified threshold, causing the instant association methods to take over the association process instead.

\noindent\textbf{Effect of different memory length.} FACT effectively utilizes all past tracking information to enhance the discriminative capability of appearance features during tracking. Table \ref{Table:7} demonstrates the impact of different memory lengths used by FAC module, ranging from 3 to the inclusion of all past tracking information. As the memory length increases, the association performance consistently improves. Using all past tracking information gives the highest association results.

\noindent\textbf{Application on Other Trackers Ablation.}
We evaluated the FACT framework on four different state-of-the-art trackers: FairMOT \cite{zhang2021fairmot}, Bot-SORT \cite{aharon2022bot}, JDE \cite{wang2020towards}, and CSTrack \cite{liang2022rethinking}. We assessed the effectiveness of the FACT framework by comparing the performance of the trackers with and without its integration. The comparative results are detailed in Table \ref{Table: different tracker}. All these trackers show approximately a 1.0 gain in IDF1 and around a 0.7 gain in HOTA, while the MOTA score remains nearly unchanged. The improvement across all methods validates the framework's compatibility, indicating that it can be easily integrated with various trackers, thereby highlighting its broad applicability.

\begin{table}[tbp]
\begin{center}
\caption{Comparison on different memory lengths.} 
\label{Table:7}
\begin{tabular}{c|cccc}
\hline
\textbf{Memory Length} & \textbf{HOTA} & \textbf{IDF1} & \textbf{MOTA} & \textbf{IDSW} \\ \hline
3                      &69.2               & 81.8          & \textbf{78.4}          & 157           \\
5                      &69.3               & 82.0          & \textbf{78.4}          & 156           \\
10                     &69.3               & 82.1          & 78.3          & 158           \\
20                     &69.5               & 82.2          & 78.3          & \textbf{155}           \\
All                    & \textbf{69.9}          & \textbf{82.7}          & 78.3          & 165           \\ \hline
\end{tabular}
\end{center}
\vspace{-15pt} 
\end{table}

\begin{table}[]
\begin{center}
\caption{Results of integrating FACT framework with 4 different state-of-the-art trackers on the MOT17 validation set.}
\label{Table: different tracker}
\begin{tabular}{c|c|llll}
\hline
\textbf{Tracker}                                 & \multicolumn{1}{l|}{\textbf{FACT}} & \textbf{HOTA} & \textbf{IDF1} & \textbf{MOTA} & \textbf{FPS} \\ \hline
\multirow{2}{*}{FairMOT \cite{zhang2021fairmot}}   &                                    & 57.3                     & 72.9                             & 69.2                                & 26.1                             \\
                          & \checkmark           &\textbf{58.1 \textcolor{darkgreen}{(+0.8)}}      & \textbf{74.0 \textcolor{darkgreen}{(+1.1)}}                              & 69.2               & 22.5                      \\ \hline
\multirow{2}{*}{Bot-SORT \cite{aharon2022bot}}  &                                       & 69.2              & 81.8                               & 78.4                             & 8.9                               \\
                          & \checkmark                                                  & \textbf{69.9 \textcolor{darkgreen}{(+0.7)}}                  & \textbf{82.7 \textcolor{darkgreen}{(+0.9)}}                              & 78.3               & 7.1                      \\ \hline 
\multirow{2}{*}{JDE \cite{wang2020towards}}      &                                      &  35.5                                         & 51.2                               & 43.6                              & 23.7                      \\
                          & \checkmark                                                  &  \textbf{40.0 \textcolor{darkgreen}{(+0.5)}}                                         & \textbf{52.3 \textcolor{darkgreen}{(+1.0)}}                              & 43.6               & 20.5                               \\ \hline
\multirow{2}{*}{CSTrack \cite{liang2022rethinking}}  &                                                  & 55.7                          & 68.8                              &  62.2                              & 18.6                               \\
                          & \checkmark                                                  &            \textbf{56.3 \textcolor{darkgreen}{(+0.7)}}                               & \textbf{69.7 \textcolor{darkgreen}{(+0.9)}}                              & 62.2               & 16.1                      \\ \hline
\end{tabular}
\end{center}
\vspace{-10pt} 
\end{table}

\noindent\textbf{Association Components Ablation.} In this section, we explore the impact of different major components within the FACT on overall tracking performance. As detailed in Table \ref{Table:2}, all components significantly contribute to metric improvements. Using IoU only as the baseline, integrating cosine distance association, CMC, and the FAC module results in a 14.1 increase in HOTA, a 25.5 increase in IDF1, and a 7.3 increase in MOTA. Among these components, cosine distance and CMC enhance both association and detection performance. This is reflected in the metrics, with improvements in HOTA by 12.5 and 0.9, respectively. Our proposed FAC module, in contrast, primarily focuses on improving the association aspect. As a result, the FAC module increases the IDF1 score by 0.9 and HOTA by 0.7. The consistent increases in HOTA and IDF1 scores for each component highlight the superior association performance of our method.\vspace{-10pt}

\begin{table}[tbp]
\begin{center}
\caption{Ablation study of different association components in the FACT. "IoU" is short for IoU distance association, "Cosine" is short for cosine distance association, "CMC" is short for camera motion compensation, and "FAC" is short for our affinity estimation with FAC module.} 
\label{Table:2}
\begin{tabular}{cccccccc}
\hline
\multicolumn{1}{l}{\textbf{IoU}} & \multicolumn{1}{l}{\textbf{Cosine}}  & \multicolumn{1}{l}{\textbf{CMC}} & \multicolumn{1}{l}{\textbf{FAC}} & \multicolumn{1}{l}{\textbf{HOTA}} & \multicolumn{1}{l}{\textbf{IDF1}} & \multicolumn{1}{l}{\textbf{MOTA}} & \multicolumn{1}{l}{\textbf{IDSW}} \\ \hline
                   \checkmark     &           &                                    &                                       & 55.8                                & 60.2                              & 71.0                              & 836                               \\
                    \checkmark    &\checkmark           &                                  &                                       & 68.3                                & 79.5                              & 77.9                              & 307                              \\
                    \checkmark    &\checkmark           &                             \checkmark       &                                       & 69.2                                  & 81.8                              & \textbf{78.4}                              & \textbf{157}                                 \\ 
                     \checkmark     & \checkmark         &                             \checkmark       &                                     \checkmark  &  \textbf{69.9}                                 &  \textbf{82.7}                                 & 78.3                                  & 165                                  \\\hline
\end{tabular}
\end{center}
\vspace{-15pt} 
\end{table}

\begin{table*}[tbp]
\begin{center}
\caption{Comparison with state-of-the-art MOT methods on the MOT17 test set.}
\label{Table:3}
\begin{tabular}{c|l|c|c|c|c|c|c|c}
\hline
Mode                      & Method         & Ref.       & HOTA (↑)                    & IDF1 (↑)                    & MOTA (↑)                    & AssA (↑)                    & DetA (↑)                    & IDSW (↓)                                        \\ \hline
offline                   & SHUSHI \cite{cetintas2023sushi}        & CVPR'2023  & 66.5                        & 83.1                        & 81.1                        & 67.8                        & 65.5                        & 1,149                                              \\ \hline
                          & SORT \cite{wojke2017simple}          & ICIP'2016  & 34.0                        & 39.8                        & 43.1                        & 31.8                        & 37.0                        & 4,852                                    \\
                          & DMAN \cite{zhu2018online}           & ECCV'2018   & -                        & 55.7                        & 48.2                       & -                        & -                       & 2,194                                               \\
                          & DAN \cite{sun2019deep}           & TPAMI'2019 & 39.3                        & 49.5                        & 52.4                        & 36.2                        & 43.1                        & 8,431                                               \\
                          & TubeTK \cite{pang2020tubetk}         & CVPR'2020  & 48.0                        & 58.6                        & 63.0                        & 45.1                        & 51.4                        & 4,137                                               \\
                          & CenterTrack \cite{zhou2020tracking}    & ECCV'2020  & 52.2                        & 64.7                        & 67.8                        & 51.0                        & 53.8                        & 3,039                                               \\
                          & SOTMOT \cite{zheng2021improving}     & CVPR'2021 & -                        & 71.9                        & 71.0                        & -                        & -                        & 5,184                                              \\
                          & PermaTrack \cite{tokmakov2021learning}     & ICCV'2021  & 55.5                        & 68.9                        & 73.8                        & 53.1                        & 58.5                        & 3,699                                              \\
                          & TrackFormer \cite{meinhardt2022trackformer}    & CVPR'2022  & 57.3                        & 68.0                        & 74.1                        & 54.1                        & 60.9                        & 2,829                                               \\
                          & CSTrack \cite{liang2022rethinking}       & TIP'2022   & 59.3                        & 72.6                        & 74.9                        & 57.9                        & 61.1                        & 3,567                                              \\
                          & FairMOT \cite{zhang2021fairmot}       & IJCV'2021  & 59.3                        & 72.3                        & 73.7                        & 58.0                        & 60.9                        & 3,303                                              \\
                          & MeMOT \cite{cai2022memot}    & CVPR'2022  & 56.9                        & 69.0                        & 72.5                        & 55.2                        & -                        & 2,724                                              \\
                         & RelationTrack \cite{yu2022relationtrack}  & TMM'2022   & 61.0                        & 74.7                        & 73.8                        & 61.5                        & 60.6                        & 1,374                                               \\
                          & MOTR \cite{zeng2022motr}           & ECCV'2022  & 62.0                        & 75.0                        & 78.6                        & 60.6                        & 63.8                        & 2,619                                               \\
                          & ByteTrack \cite{zhang2022bytetrack}      & ECCV'2022  & 63.1                        & 77.3                        & 80.3                        & 62.0                        & 64.5                        & 2,196                                              \\
                          & OC-SORT \cite{cao2023observation}       & CVPR'2023  & 63.2                        & 77.5                        & 78.0                        & 63.2                        & -                           & 1,950                                              \\
                          & StrongSORT \cite{du2023strongsort}    & TMM'2023   & 63.5                        & 78.5                        & 78.3                        & 63.7                        & 63.6                        & 1,446                                               \\
 \multirow{-11}{*}{online}                          & BoT-SORT-ReID \cite{aharon2022bot} & arxiv'2022 &  65.0 &  80.2 &  80.5 &  65.5 &  64.9 &  1,212  \\
                          & MotionTrack \cite{qin2023motiontrack}    & CVPR'2023  & 65.1                        & 80.1                        & \textbf{81.1}               & 65.1                        & \textbf{65.4}               & 1,140                                     \\
                          & BoostTrack+ \cite{stanojevic2024boosttrack}             & MVA'2024 &  66.4 &  81.8 &  80.6 &  67.7 &  65.4 & 1,086  \\
& FACT     & ours       & 65.3               & 80.5               & 80.4                        & 65.9               & 65.0                        & 1,367                                                \\ 
 & FACT+                  & ours       & \textbf{66.8}               & \textbf{82.9}               & 80.4                        & \textbf{68.7}               & 65.3               & \textbf{1,026}                                                  \\ \hline
\end{tabular}
\end{center}
\vspace{-10pt} 
\end{table*}

\begin{table*}[tbp]
\begin{center}
\caption{Comparison with state-of-the-art MOT methods on the MOT20 test set.}
\label{Table:4}
\begin{tabular}{c|l|c|c|c|c|c|c|c}
\hline
Mode                      & \multicolumn{1}{c|}{Method} & Ref.       & HOTA (↑)                    & IDF1 (↑)                    & MOTA (↑)                    & AssA (↑)                    & DetA (↑)                    & IDSW (↓)                                         \\ \hline
offline                   & SHUSHI \cite{cetintas2023sushi}                      & CVPR'2023  & 64.3                        & 79.8                        & 74.3                        & 67.5                        & 61.5                        & 706                                                  \\ \hline
                          & SORT \cite{wojke2017simple}                       & ICIP'2016  & 36.1                        & 45.1                        & 42.7                        & 35.9                        & 36.7                        & 4,470                                       \\

                          & CSTrack \cite{liang2022rethinking}                    & TIP'2022   & 48.5                        & 59.4                        & 65.0                        & 45.2                        & 53.3                        & 3,608                                               \\
                          & MeMOT \cite{cai2022memot}                 & CVPR'2022 & 54.1                        & 66.1                        & 63.7                        & 55.0                        & -                        & 1,938                                                \\
                          & FairMOT \cite{zhang2021fairmot}                    & IJCV'2021  & 54.6                        & 67.3                        & 61.8                        & 54.7                        & 54.7                        & 5,243                                               \\
                                                    & SOTMOT \cite{zheng2021improving}                    & CVPR'2021  & -                        & 71.4                        & 68.6                        & -                        & -                        & 4,209                                               \\
                          & RelationTrack \cite{yu2022relationtrack}              & TMM'2022   & 56.5                        & 70.5                        & 67.2                        & 56.4                        & 56.8                        & 4,243                                                \\
                          & ByteTrack \cite{zhang2022bytetrack}                  & ECCV'2022  & 60.9                        & 74.9                        & 75.7                        & 59.9                        & 62.0                        & 1,347                                               \\
                          & StrongSORT \cite{du2023strongsort}                 & TMM'2023   & 61.5                        & 75.9                        & 72.2                        & 63.2                        & 59.9                        & 1,066                                                \\
                          & OC-SORT \cite{cao2023observation}                   & CVPR'2023  & 62.1                        & 75.9                        & 75.5                        & 62.0                        & -                           & 913                                          \\
                          & MotionTrack \cite{qin2023motiontrack}                & CVPR'2023  & 62.8                        & 76.5                        & \textbf{78.0}               & 61.8                        & 64.0                        & 1,165                                               \\
                          & BoT-SORT-ReID \cite{aharon2022bot}             & arxiv'2022 &  63.3 &  77.5 &  77.8 &  62.9 &  64.0 & 1,313  \\
                         & BoostTrack+ \cite{stanojevic2024boosttrack}             & MVA'2024 &  66.2 &  81.5 &  77.2 &  68.6 &  64.1 & 966  \\
\multirow{-11}{*}{online}  & FACT                  & ours       & 65.0               & 79.8               & 77.9                        & 66.1               & 64.2               & 1,263                                                  \\ 
 & FACT+                  & ours       & \textbf{67.2}               & \textbf{83.6}               & 77.5                        & \textbf{70.6}               & \textbf{64.2}               & \textbf{723}                                                \\  \hline
\end{tabular}
\end{center}
\vspace{-10pt} 
\end{table*}

\begin{figure*}[]
    \centering
    \includegraphics[width=1.0\textwidth,trim=50 150 200 50,clip]{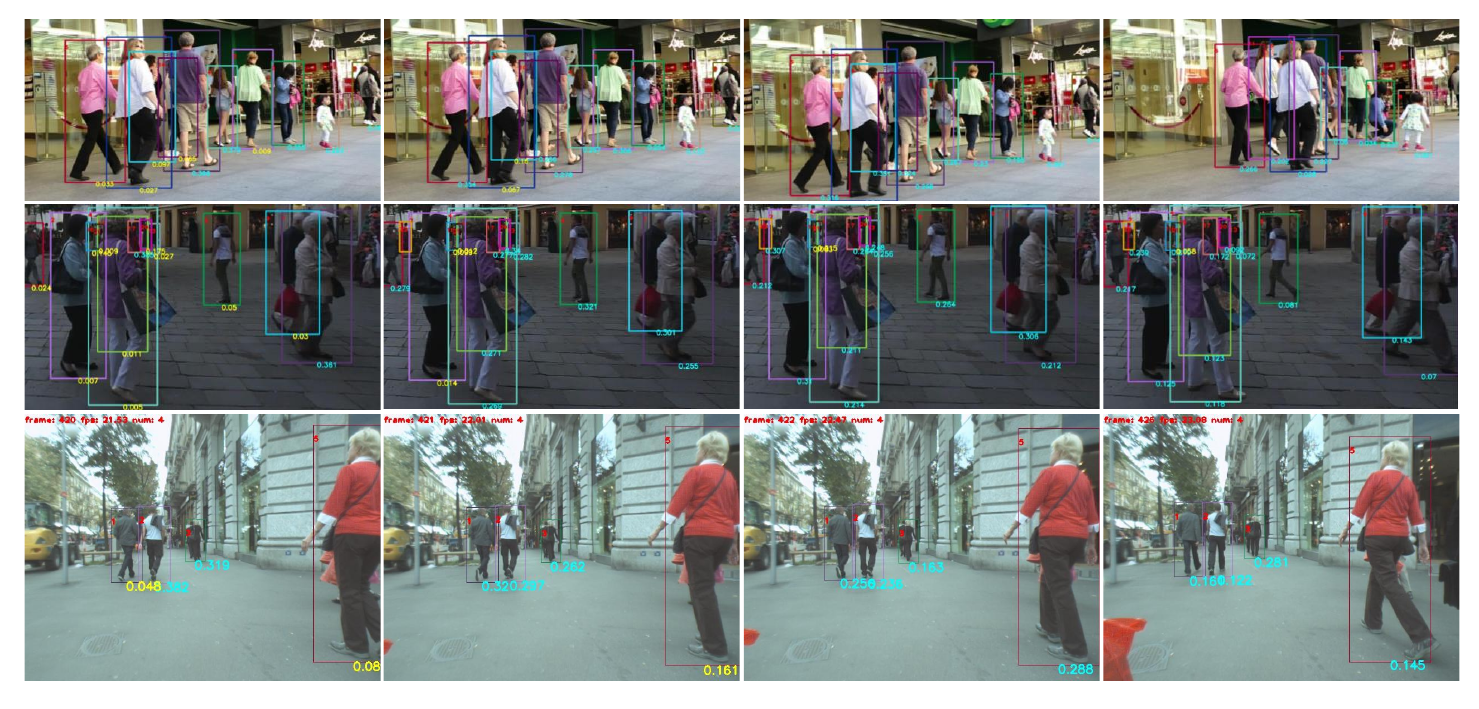}
    \caption{\textbf{Visualization of sample tracking results. We select 3 sequences from the validation set of the MOT17. The yellow value under each box represents using the cosine distance association and the corresponding results, the blue value represents using the affinity estimation and the corresponding results.} }
    \label{fig: visualization}
    \vspace{-10pt} 
\end{figure*}

\subsection{Benchmark Evaluation}

We compared FACT with the latest state-of-the-art trackers on the test sets of MOT17 and MOT20 under the private detection protocol. The results, which are directly downloaded from the official MOT challenge website, are presented in Table \ref{Table:3} and Table \ref{Table:4}, respectively. Recently, BoostTrack+ \cite{stanojevic2024boosttrack} achieved the state-of-the-art results by introducing a new instant association technique that builds on the existing cosine distance and IoU distance association techniques. Since BoostTrack+ is orthogonal to ours, we created FACT+ with its new instant association technique and post-processing method. This integration introduces only minimal computational overhead compared to the original FACT. The FPS for FACT+ on the MOT17 and MOT20 test datasets is 12.3 and 3.2, respectively, on a 2080ti GPU.

\noindent\textbf{Real-Time Testing.} Following most of the state-of-the-art methods \cite{qin2023motiontrack, stanojevic2024boosttrack}, we evaluate the real-time performance of FACT+ using a current commercial off-the-shelf RTX4090 GPU. On this modern hardware, FACT+ achieves 21.2 FPS on the MOT17 test dataset, validating its capability for real-time tracking. Additionally, we analyze the real-time performance of the proposed method with different trackers in Table \ref{Table: different tracker}. As demonstrated, the proposed method can be applied to different trackers with minimal extra time cost, it can generally achieve real-time performance, provided the trackers themselves are capable of doing so.\vspace{-5pt}




\vspace{\baselineskip}
\noindent\textbf{MOT17.} In the MOT17 assessment, both FACT and FACT+ achieve state-of-the-art performance. Notably, FACT+ outperforms all online and offline trackers in HOTA, IDF1, AssA, and IDSW. Our methods excel due to their ability to adapt to occlusions through learning from tracking and improved discriminative capability by utilizing all past features for learning. As a result, FACT+ surpasses BoostTrack+ and Bot-SORT-ReID, which uses the same feature extractor, by 0.5 and 1.8 in HOTA, 1.1 and 2.7 in IDF1, 1.0 and 3.2 in AssA. FACT+ also demonstrates excellent IDSW among all methods, indicating its effectiveness in solving target association under occlusions through learning from tracking. There is no improvement in MOTA, largely because our method's focus on ReID feature association does not significantly impact detection performance. The superior performance of FACT and FACT+ on association-related metrics comprehensively validates their effectiveness. \vspace{-5pt} 


\vspace{\baselineskip}


\noindent\textbf{MOT20.} Compared to MOT17, MOT20 presents a more challenging environment due to its crowded scenes, which frequently result in track interruptions from occlusions. Despite these challenges, our FACT+ ranks in the HOTA metric on the MOT20 leaderboard among all available public methods (both online and offline). Notably, FACT already achieves outstanding tracking accuracy even with naive cosine distance and IoU association methods (65.0 HOTA, 79.8 IDF1, etc.). With enhanced instant association techniques, FACT+'s performance improves further (67.2 HOTA, 83.6 IDF1, etc.), surpassing BoostTrack+ and Bot-SORT-ReID by 1.0 and 3.6 in HOTA, and by 2.1 and 6.1 in IDF1. It even outperforms the offline state-of-the-art method SHUSHI by a large margin of 2.6 in HOTA. Given the complexity of MOT20's scenarios, FACT's superior tracking performance emphasizes the importance of our continual learning methodology for the MOT task. \vspace{-5pt}



\subsection{Visualization Results}

Visual results of FACT are illustrated in Fig. \ref{fig: visualization}, sourced from the MOT17 validation set. These images primarily highlight FACT's capabilities under varying environmental complexities. Different colors represent the various association strategies used for each target. Yellow denotes targets associated using the cosine distance association, while blue indicates targets associated through the affinity estimation of the FAC module (affinity distance). It's evident that the FAC module takes a few frames to initialize. Following this initialization, the confidence of the affinity estimation improves with the introduction of more training samples. \vspace{-5pt} 


\section{Conclusion}
In this paper, we present an innovative framework for MOT, known as FACT, which enables real-time tracking and target feature learning by leveraging all past tracking information. Additionally, we show that the framework can be integrated with various state-of-the-art feature-based trackers, thereby improving their tracking ability with minimal extra computation. To achieve this, we designed the FAC module, a neural network that can be trained online using all past tracking information simultaneously. To minimize the training time for real-time performance, we propose an analytic learning-based methodology to achieve this training recursively. We also demonstrate through experiments that the FAC module can be easily integrated with several state-of-the-art appearance-based trackers to improve tracking performance with moderate time cost. Additionally, we designed a two-stage association module to enhance the FAC module-based affinity estimation with instant association capability, ensuring robust tracking performance for new tracks. Benchmark experiments validate FACT's superior efficacy, showing that it outperforms many state-of-the-art methods on the MOT17 and MOT20 benchmarks.\vspace{-5pt}

\ifCLASSOPTIONcaptionsoff
  \newpage
\fi

\bibliographystyle{IEEEtran}

\end{CJK}
\end{document}